\pdfoutput=1

\documentclass[11pt]{article}
\usepackage{widetable}
\usepackage{multirow}

\usepackage{acl}
\usepackage{gb4e}
\noautomath

\usepackage{times}
\usepackage{latexsym}

\usepackage[T1]{fontenc}

\usepackage[utf8]{inputenc}

\usepackage{microtype}

\usepackage{inconsolata}
\usepackage{ifthen}
\newboolean{DRAFT}
\setboolean{DRAFT}{true}
\ifthenelse{\boolean{DRAFT}}
{
\definecolor{mediumblue}{rgb}{0.0, 0.0, 0.8}
\definecolor{mediumred-violet}{rgb}{0.73, 0.2, 0.52}
\definecolor{mediumcandyapplered}{rgb}{0.89, 0.02, 0.17}
\newcommand{\kot}[1]{{\color{black} {#1}}}
\newcommand{\js}[1]{{\color{mediumred-violet} JS: {#1}}}
\newcommand{\ryo}[1]{{\color{mediumblue} RYO: {#1}}}
\newcommand{\dkw}[1]{{\color{mediumcandyapplered} DKW: {#1}}}
}{
\newcommand{\kot}[1]{}
\newcommand{\js}[1]{}
\newcommand{\ryo}[1]{}
\newcommand{\dkw}[1]{}
}

\title{Verifying Claims About Metaphors\\ with Large-Scale Automatic Metaphor Identification}

\author{Kotaro Aono \ \ \ \ Ryohei Sasano \ \ \ \  Koichi Takeda\\
  Graduate School of Informatics, Nagoya University \\
  \texttt{aono.kotaro.i1@s.mail.nagoya-u.ac.jp},
  \texttt{\{sasano,takedasu\}@i.nagoya-u.ac.jp}\\}

\begin{document}

\maketitle
\begin{abstract}
There are several linguistic claims about \kot{situations where words} are more likely to be used as metaphors.
However, few studies have sought to verify such claims with large corpora.
This study entails a large-scale, corpus-based analysis of certain existing claims about verb metaphors, by applying metaphor detection to sentences extracted from Common Crawl and using the statistics obtained from the results.
The verification results indicate that the direct objects of verbs used as metaphors tend to have lower degrees of concreteness, imageability, and familiarity, and that metaphors are more likely to be used in emotional and subjective sentences.
\end{abstract}

\section{Introduction}
A metaphor is a figurative expression without a direct simile. 
Such expressions appear frequently in all kinds of documents, and there have been numerous studies on metaphors.
In the field of cognitive linguistics, the most representative work is that of \citet{mwlb}.
They emphasize that metaphor is not just a rhetorical device, but a significant function that largely reflects human cognition.
Consider the following examples.
\begin{exe}
 \ex He \textit{attacked} weak points in my argument.\label{ex:attack}
 \ex You can't \textit{win} this argument.\label{ex:win}
\end{exe}
In these \kot{examples}, terms associated with \textsc{war}, such as \textit{attack} and \textit{win} are used in relation to the concept of \textsc{argument}.
Such metaphors are used because the concept of \textsc{argument} includes elements of winning and losing, as well as the strategic use of tactics for both offense and defense. By using metaphor, one can understand the abstract concept of \textsc{argument} by relating it to the more concrete concept of \textsc{war}.

Lakoff and Johnson contended that the fundamental nature of metaphor lies in the understanding of one concept through another, and they asserted that metaphor is at the core of human cognition.
They called this cognitive function ``conceptual metaphor,'' which has had significant influence in the field of linguistics.
As a result, metaphor has emerged as a crucial topic within cognitive linguistics, giving rise to numerous claims. 
However, most existing studies discuss each claim in terms of a small number of examples, whereas few such studies verify claims by analyzing a large corpus.
Hence, this study attempts to verify certain claims about metaphors by leveraging natural language processing techniques and a large corpus.

\begin{table*}[t]
\small
\centering
\renewcommand{\arraystretch}{1.4}
\begin{tabular}{l}
\hline
\textbf{Claims about direct objects in verb metaphors~\cite{japmet}}\\
\hline
\textbf{A:} The direct objects in metaphorical examples have  less concreteness than those in literal examples.\\
\textbf{B:} The direct objects in metaphorical examples have less imageability than those in literal examples.\\
\textbf{C:} The direct objects in metaphorical examples have  less familiarity than those in literal examples.\\
\hline
\textbf{Claims about relations between metaphor and emotion/subjectivity~\cite{roleofmet}}\\
\hline
\textbf{D:} Sentences with an emotional polarity have a high metaphor usage rate.\\
\textbf{E:} Sentences describing subjective experience have a high metaphor usage rate.\\
\hline
\end{tabular} 
\caption{Claims about metaphors to be verified in this study.}
\label{object-claims}
\vspace{-0.8ex}
\end{table*}

\section{Claims about Metaphors}
Conceptual metaphor~\cite{mwlb} is a cognitive process in which one concept is understood in terms of another concept through the use of metaphor.
In the preceding examples, \textsc{argument} is perceived via a projection on a concept like \textsc{war}, in which there are winners, losers, and strategies.
In this metaphorical projection, a concept like \textsc{war}, which serves as the projection's source, is called the source domain, while a concept like \textsc{argument}, which represents the projection target, is called the target domain.

\citet{japmet} argues that the source domain is concrete, is easily described, and involves concrete experiences, whereas the target domain is abstract, is difficult to describe, and involves less physically concrete experiences.
For example, while \textit{the argument} in (\ref{ex:win}) and \textit{the battle} in (\ref{ex:battle}) are both direct objects of the verb \textit{win}, \textit{argument}, which is used as a metaphor, is less concrete, less imageable, and less familiar.  
Such properties have already been exploited for metaphor identification in natural language processing~\cite{turney,tsvetkov,rai}.

\begin{exe}
 \ex You can't \textit{win} this battle.\label{ex:battle}
\end{exe}

\citet{roleofmet} argue that ``metaphors can paint a richer and more detailed picture of our subjective experience than can be expressed by literal language.'' They further note that ``the metaphorical description does represent an attempt to characterize the quality of a subjective state.''
\citet{LEDOUX201867} argue that ``subjective emotional experience is the essence of emotion, and objective manifestations in behavior, body, or brain physiology are at best indirect indicators of these internal experiences.''
If these claims are true, then sentences that describe subjective experiences and express emotions are more likely to contain metaphors.

In this study, regarding the above claims by \citet{japmet} and \citet{roleofmet}, we leveraged a large corpus and sought to verify five claims about metaphors which are listed in Table \ref{object-claims}.
Of these claims, Claims A, B, and C are about conceptual metaphors, while Claims D and E are about \kot{emotion and subjectivity of metaphorical expressions}.

\section{Preparation for Verifying Claims}
\subsection{Metaphor Identification Model}
\label{Section::Classifier}
In this study, we used MisNet~\cite{misnet} for metaphor identification.
MisNet incorporates the two concepts of the metaphor identification procedure (MIP)~\cite{mip} and selectional preference violation (SPV)~\cite{spv} into metaphor identification.
The former concept is based on the idea that word's meaning varies between when it is used metaphorically and when it is used literally.
A metaphorical word can be identified by considering the gap between word embeddings in a basic usage context and word embeddings in the given context.
The latter concept, SPV, is based on the idea that whether a word is a metaphor can be determined by its semantic difference from surrounding words; thus, SPV focuses on the difference between the target word's embedding in the sentence vector and the context.

We used a model\footnote{\url{https://github.com/SilasTHU/MisNet}} trained on the VU Amsterdam Metaphor Corpus~\cite{vuamc}, a dataset of 190,000 lexical units (tokens) that were manually annotated with metaphorical and non-metaphorical labels. The annotations are based on MIPVU, a metaphor identification method that refines MIP. 

The performance of the metaphor identification model on a data that was not used for training is provided in Appendix \ref{PEM}.

\subsection{Concreteness, Imageability, Familiarity}
We used a dataset\footnote{\url{https://github.com/clarinsi/megahr-crossling/}} created by \textbf{L}jube\v{s}i\'{c}, \textbf{F}i\v{s}er, and \textbf{P}eti-Stanti\'{c}~\shortcite{objs-dataset-conc} to evaluate the concreteness and imageability of words. 
We refer to this as the \textbf{LFP} dataset. It comprises words with concreteness and imageability scores and was obtained by applying models to 77 languages. 
The models were trained by SVM regression models and feed-forward networks using English and Croatian training data. 
The English portion of the LFP dataset comprises approximately 100,000 English words, with assigned concreteness and imageability scores.
The scores range from 0.87 to 5.35 for concreteness and from 1.77 to 5.26 for imageability.

To evaluate familiarity, we used a dataset on the complexity of words: \textbf{W}ord-\textbf{C}omplexity \textbf{L}exico\footnote{\url{https://github.com/mounicam/lexical_simplification}}~\cite{objs-dataset-comp} or \textbf{WCL}.
The WCL dataset comprises 15,000 English words annotated with  a 6-point word complexity score.
The annotations were created by 11 non-native but fluent English speakers who rated each word on a scale, ranging from very simple (1) to very complex (6). 
For each word, the word complexity score is the average of the annotator's ratings.
Thus, each word complexity score $c$ in the WCL dataset ranges from 1 to 6; here, we used $6-c$ as the familiarity score. 

\subsection{Example Collection}
To verify the claims about verb metaphors, we collected examples of verb-object pairs.
First, we extracted sentences containing verb-object pairs from a pool of 1 billion English sentences obtained from the CC-100\footnote{https://data.statmt.org/CC-100} corpus. 
We then input each sentence to the metaphor identification model to determine whether the sentence's verbs were metaphorical or literal.
After lemmatizing, we treated each verb-object pair as a metaphor example if more than 70\% of the occurrences of the pair in the actual text are determined to be metaphorical. 
Conversely, if less than 30\% of the occurrences in the actual text are determined to be metaphorical, we treated it as a non-metaphorical example.
For example, we treated the pair \textit{attack-idea} as a metaphorical example because 99\% of \textit{attacks} that appeared with \textit{idea} in the actual text were determined to be metaphorical, whereas we treated the pair \textit{attack-ship} as a non-metaphorical example because only 18\% of \textit{attacks} that appeared with \textit{ship} were determined to be metaphorical.

We focused our analysis on the 49 verbs listed in Table \ref{tab:allverbs}, for which abundant examples could be collected.\footnote{The detailed selection procedure is given in Appendix~\ref{verbs-collection}.}
Specifically, we collected sentences containing these verbs from the 1 billion CC-100 sentences.
In counting metaphorical and non-metaphorical examples, we excluded objects that were not included in the data used to determine concreteness, imageability, and familiarity. 
As a result, approximately 80\% of the total objects were used.
The maximum, minimum, and median numbers of different verb-object pairs for each verb were respectively 2377, 13, and 743, for the metaphorical cases, and 3292, 18, and 679, for the literal cases.
\begin{table}[t]
\small
\centering
\begin{tabular}{lllll}
\hline
pocket & buy     & eat     & pull  & build   \\
exchange  & spell   & lift    & join & piece    \\
allow  & milk    & gain & pick & break  \\
welcome   & tell    & view    & kiss & save    \\
attack & plant   & make & watch & track   \\
witness   & meet    & ride    & find & raise   \\
express   & kill    & carry   & voice & shed   \\
cross  & hand    & free    & cut & harm   \\
hold   & waste   & send    & lose & take     \\
raid   & put     & cost    & teach \\
\hline
\end{tabular}
\caption{List of 49 verbs used in the analysis.}
\label{tab:allverbs}
\end{table}

For claims D and E, we collected sentences with different emotional polarity and subjectivity.
First, we applied Stanza's sentiment analysis model~\cite{stanza} to classify each sentence into one of three types: positive, neutral, and negative.
\kot{
The performance of the sentiment classification model on a data that was not used for training is provided in Appendix \ref{PEM}.
}
Then, for subjectivity, we treated sentences in which the subject was a first person pronoun (i.e., \textit{I, we}) as likely to be subjective, and sentences in which the subject was a third person pronoun (i.e., \textit{he, she, they}) as likely to be objective.
We collected a total of 120,000 sentences from the CC-100 corpus, comprising 20,000 sentences for each combination of the three emotional polarity types and two subjectivity types.
To remove bias due to sentence length, we collected the sentences so as to ensure that the distribution of their lengths was consistent across the six groups.
Then, we applied the metaphor identification model to the collected sentences. 
We considered the sentence as containing a metaphor if even one word in the sentence was determined to be metaphorical, and calculated the percentage of sentences that contain a metaphor.
By comparing the metaphor usage rates across groups, we analyzed the relationship between emotional polarity/subjectivity and the metaphor usage rate.

\section{Experiments}
\begin{table*}[]
\small
\centering
\begin{tabular}{l|lll}
\hline
Verb                     & Concreteness [CI]           & Imageability [CI]           & Familiarity [CI]           \\ \hline
\\[-6pt]
\multirow{2}{*}{buy}     & 2.50 [2.16, 2.83] & 3.21  [2.93, 3.50]  & 3.31  [2.79, 3.83] \\
                         & 3.86         [3.35, 4.15] & 3.80         [3.78, 4.40] & 4.11        [3.53, 4.32] \\[4pt] 
\multirow{2}{*}{break}   & 3.25         [3.20, 3.34] & 3.54         [3.57, 3.69] & 3.68        [3.47, 3.77] \\
                         & 4.35         [4.31, 4.40] & 4.38         [4.38, 4.45] & 4.00        [3.90, 4.10] \\[4pt] 
\multirow{2}{*}{send}    & 3.24         [3.35, 3.53] & 3.76         [3.76, 3.90]  & 3.76        [3.65, 3.87] \\
                         & 3.70         [3.54, 3.70] & 4.03         [3.80, 3.94] & 3.99        [3.74, 3.96] \\[4pt] 
\multirow{2}{*}{welcome} & 3.39         [3.08, 3.45] & 3.70         [3.50, 3.78] & 3.73        [3.54, 3.81] \\
                         & 3.78         [3.66, 3.90] & 4.08         [3.98, 4.18] & 3.45        [3.34, 3.56] \\[2pt] \hline
                         \\[-6pt]
\multirow{2}{*}{All verbs}   & 3.16         [3.17, 3.18] & 3.57         [3.57, 3.58] & 3.63        [3.61, 3.64] \\
                         & 3.99         [4.02, 4.03] & 4.18         [4.20, 4.21] & 3.87        [3.87, 3.89] \\[2pt] \hline
\end{tabular}
\caption{\kot{Average scores} with 95\% Confidence intervals (CI) of concreteness, imageability, and familiarity for objects used with four specific verbs and with all 49 verbs in \kot{total}.
\kot{The upper and lower numbers for each verb} indicate the average score and confidence intervals when the verb is used as a metaphor \kot{and} non-metaphor, respectively.}
\label{tab:all-result}
\end{table*}

\subsection{Concreteness of Objects}
To verify Claim A, we calculated the average concreteness scores for both metaphorical and non-metaphorical examples for each verb.
Table~\ref{tab:all-result} shows the results for four specific verbs and for the verb in total as well as the imageability and familiarity, as explained below.
This result indicates that direct objects in metaphorical examples tend to be less concrete than direct objects in non-metaphorical examples, which is consistent with Nabeshima's claim~\cite{japmet} that the target domain's concreteness is lower in conceptual metaphors~(Claim A). 
Table~\ref{tab:conc-table} in Appendix~\ref{Verification-detail} gives the detailed results, including the metaphor usage rate for all 49 verb, the average concreteness and the object examples.
We can see that for all 49 verbs, the average direct object concreteness score in the metaphorical examples is lower than that in the non-metaphorical examples.
The number of verbs that agree with the claim A is 49 out of 49, and when this distribution is tested with a binomial test, we can see a statistically significant bias (p < 0.0001).

\subsection{Imageability of Objects}
Next, to verify Claim B, we calculated the average imageability scores for both metaphorical and non-metaphorical examples for each verb.
Table~\ref{tab:all-result} also shows the results.
As with concreteness, this result indicates that direct objects in metaphorical examples tend to be less imageable than direct objects in non-metaphorical examples, which is consistent with Nabeshima's claim~\cite{japmet} that the target domain's imageability is lower in conceptual metaphors~(Claim B). 
Table~\ref{tab:img-table} in Appendix~\ref{Verification-detail} gives the detailed results, including the metaphor usage rates for all 49 verbs, the average imageability and the object examples.
We can see that for all 49 verbs, the average direct object imageability score in the metaphorical examples is lower than that in the non-metaphorical examples.
The number of verbs that agree with the claim A is 49 out of 49, and when this distribution is tested with a binomial test, we can see a statistically significant bias (p < 0.0001).
\subsection{Familiarity of Objects}
Then, to verify Claim C, we calculated the average familiarity scores for both the metaphorical and non-metaphorical examples for each verb.
Table~\ref{tab:all-result} also includes the results.
This result indicates that direct objects in metaphorical examples tend to be less familiar than direct objects in non-metaphorical examples, which is consistent with Nabeshima's claim~\cite{japmet} that the target domain's familiarity is lower in conceptual metaphors~(Claim C). 
Table~\ref{tab:fam-table} in Appendix~\ref{Verification-detail} gives the detailed results, including the metaphor usage rates for all 49 verbs, the average familiarity and the object examples. 
In this case, for 45 out of the 49 verbs, excluding \textit{welcome}, \textit{attack}, \textit{kiss}, and \textit{kill}, the average direct object familiarity score in the metaphorical examples is lower than that in the non-metaphorical examples. 
The number of verbs that agree with the claim A is \kot{45} out of 49, and when this distribution is tested with a binomial test, we can see a statistically significant bias (p < 0.0001).
\begin{table}[t]
\small
\centering
\begin{tabular}{c|cc}
\hline
Emotion \textbackslash \ Subject    & 1st person  & 3rd person \\
\hline
Positive & 0.896 & 0.893 \\
Neutral  & 0.868 & 0.857 \\
Negative & 0.883 & 0.866\\
\hline
\end{tabular}
\caption{
Metaphor usage rates for groups categorized by the emotion and subject. \kot{If claim D is true, then the value should be smaller in the neutral case than in the other cases. If claim E is true, then the value should be larger in the 1st person case than in the other cases.  }}
\label{tab:Iwemetper}
\end{table}

\begin{table}[t]
\small
\centering
\renewcommand\arraystretch{1.2}
\setlength{\tabcolsep}{2cm}
\begin{tabular}{@{ \ }c@{ \ }|@{ \ }c@{ \ \ }c@{ \ \ \ }c@{ \ \ }c@{ \ }}
\hline
Group & \hspace{1ex}Samples\hspace{1ex} & MUR & Diff & P-value    \\
\hline
\begin{tabular}[l]{@{}l@{}}Neutral\\ Otherwise\end{tabular} 
& \begin{tabular}[c]{@{}c@{}}40000\\ 80000\end{tabular}
& \begin{tabular}[c]{@{}c@{}}0.862\\ 0.885\end{tabular} 
& -0.023 & $<$0.0001 \\
\hline
\begin{tabular}[l]{@{}l@{}}Positive\\ Otherwise\end{tabular}
& \begin{tabular}[c]{@{}c@{}}40000\\ 80000\end{tabular}  
& \begin{tabular}[c]{@{}c@{}}0.895\\ 0.868\end{tabular}
& \ 0.027 & $<$0.0001 \\
\hline
\begin{tabular}[l]{@{}l@{}}Negative\\Otherwise\end{tabular} 
& \begin{tabular}[c]{@{}c@{}}40000\\ 80000\end{tabular}  
& \begin{tabular}[c]{@{}c@{}}0.874\\ 0.879\end{tabular} 
& -0.005 & 0.0337\\
\hline
\begin{tabular}[l]{@{}l@{}}First person\\Third person\end{tabular} 
& \begin{tabular}[c]{@{}c@{}}60000 \\ 60000\end{tabular} 
& \begin{tabular}[c]{@{}c@{}}0.882\\ 0.871\end{tabular} 
& \ 0.011 & $<$0.0001 \\
\hline
\end{tabular}
\caption{Verification results for the mean metaphor usage rate (MUR) in group categorized by the emotion and subject.}
\label{tab:Iwepermtest}
\end{table}

\subsection{Emotion/Subjectivity-Metaphor Usage Rate Relationship}
Finally, for our analysis to verify Claims D and E, Table \ref{tab:Iwemetper} lists the metaphor usage rates for each group. Moreover, Table \ref{tab:Iwepermtest} summarizes the differences in metaphor usage rates and the results of a permutation test\footnote{The p-values were estimated by randomly generating 100,000 permutations. A significance level of 0.01 was used, and the Bonferroni method was applied to address the multiple-comparisons problem.} 
for the following pairs: neutral/otherwise, positive/otherwise, negative/otherwise, and first person/third person.
First, regarding emotional polarity, we found that sentences with emotional polarity show significantly higher metaphor usage rates than neutral sentences. Furthermore, positive sentences show significantly higher metaphor rates than neutral or negative sentences.
This result indicates that the frequency of metaphor usage was increased in sentences with emotional polarity, especially in sentences with positive polarity, thus supporting Claim D.

In addition, this result shows that sentences in which the subject was first person pronoun had significantly higher metaphor usage rates than sentences in which the subject was third person pronoun.
If the assumption that sentences with a first person pronoun as the subject are more likely to express subjectivity is true, then this result suggests that metaphors are more likely to be used in sentences expressing subjective content, thus supporting Claim E.

\section{Conclusion}
In this paper, we have sought to verify existing claims about metaphors by using a large corpus.
Specifically, we examined three claims related to conceptual metaphors and two claims related to the connections between metaphors and the strength of emotional polarity and subjectivity. 
As a result, we found that the direct objects of verbs used as metaphors tend to have lower degrees of concreteness, imageability, and familiarity, and that metaphors are more likely to be used in emotional and subjective sentences.

\section*{Limitation}
First, Regarding comprehensive analysis of the nature of metaphors in language, one potential limitation of this study is that the analysis was limited to English. 
Metaphors are used in many languages, but it is unclear whether the results obtained in this study are valid also for other languages.
Another limitation of this study is that the determination of whether or not each example is a metaphor is done automatically, and the performance of the system has not been adequately analyzed.
Second, although prior studies have shown that the performance of automatic determination is reasonably high, if there was a particular tendency toward error, it is possible that this tendency could be a bias.
Third, in this study, the examples were collected from CommonCrawl, which is a corpus of texts on the Web, and thus contains mostly written language.
It is unknown whether similar results can be obtained when analyzing examples collected from a corpus with different characteristics, such as spoken \kot{languages}.
\kot{
Finally, because we could not find a large dataset that directly examined the familiarity score, we approximated the familiarity score using the complexity score, for which a larger dataset exists. We assumed that the complexity score and the familiarity socre negatively correlated, but if this assumption is incorrect, there exists the possibility that the conclusions reached in this paper are incorrect.
}
\bibliography{custom}

\appendix

\section{Appendices}
\kot{
\subsection{Performance of each Model on Datasets that were not Used for Training}
\label{PEM}
To evaluate the performance of the models used in this study , i.e., metaphor identification model and sentiment classification model, we measured the accuracy of each model on a dataset that was not used for training.
Regarding the metaphor identification model, we evaluated the accuracy for the three datasets. We obtained an accuracy of 0.61 for the TroFi dataset~\cite{trofi1,trofi2}, 0.81 for the MoH-X dataset~\cite{moh}, and 0.80 for the TSV dataset \cite{tsvetkov}.
Regarding the sentiment classification model, we evaluated the accuracy for the two datasets. We obtained an accuracy of 0.57 for the Tweeteval dataset~\cite{tweeteval} and 0.63 for the Dynasent dataset~\cite{dynasent}.
}
\subsection{Verb Selection Procedure}
\label{verbs-collection}
Here, we explain how we choose the 49 verbs that were used for the analysis in this study.
We only considered verbs that are used primarily as transitive verbs, specifically those that were transitive in more than 70\% of the instances that we collected, because verb metaphors with direct objects were the focus of our analysis.
In addition, to ensure comparability, we only analyzed verbs for which there were more than 10 different verb-object pairs for both metaphorical and literal examples.

\subsection{Verification Details}
Here, we report the detailed verification results. For concreteness, imageability, and familiarity, Tables~\ref{tab:conc-table}, \ref{tab:img-table}, and \ref{tab:fam-table} respectively summarize the metaphor usage rates for all 49 verbs, the average of each index for both metaphorical and nonmetaphorical examples, and object examples for each usage.
\label{Verification-detail}

\begin{table*}[t!]
\small
\centering
\begin{tabular}{l|l|l|c}
\hline
Verb (Metaphor rate) & Metaphorical (Object examples) & Non-metaphorical (Object examples) & Diff\\ \hline

pocket (0.28) & 3.12 (profit: 2.68, fee: 3.53, $\cdots$) & 4.05 (money: 3.88, cash: 4.16, $\cdots$) & -0.93 \\
eat (0.04) & 3.41 (world: 3.04, space: 3.47, $\cdots$) & 3.85 (food: 3.99, meal: 4.46, $\cdots$) & -0.44 \\
pull (0.36) & 3.17 (stop: 4.15, string: 3.87, $\cdots$) & 4.17 (trigger: 3.65, hair: 4.43, $\cdots$) & -0.99 \\
spell (0.65) & 2.86 (end: 3.24, trouble: 3.23, $\cdots$) & 3.54 (name: 3.41, word: 3.20, $\cdots$) & -0.67 \\
join (0.97) & 3.28 (team: 3.36, group: 3.22, $\cdots$) & 4.02 (guy: 4.09, parent: 3.46, $\cdots$) & -0.73 \\
milk (0.76) & 3.29 (system: 3.12, money: 3.88, $\cdots$) & 4.18 (cow: 4.88, goat: 4.86, $\cdots$) & -0.89 \\
express (0.99) & 2.98 (gratitude: 2.31, concern: 2.11, $\cdots$) & 4.12 (milk: 4.64, gland: 3.94, $\cdots$) & -1.14 \\
pick (0.25) & 3.13 (fight: 3.67, pace: 3.63, $\cdots$) & 4.09 (phone: 4.48, book: 3.68, $\cdots$) & -0.96 \\
make (0.87) & 3.34 (sense: 1.28, decision: 2.53, $\cdots$) & 4.28 (own: 2.36, these: 2.37, $\cdots$) & -0.94 \\
tell (0.04) & 3.28 (difference: 2.40, reality: 2.23, $\cdots$) & 3.47 (truth: 2.48, that: 2.57, $\cdots$) & -0.19 \\
view (0.35) & 3.04 (world: 3.04, thing: 2.85, $\cdots$) & 3.63 (video: 3.64, list: 3.20, $\cdots$) & -0.59 \\
kiss (0.05) & 3.59 (sky: 4.19, chance: 2.85, $\cdots$) & 3.86 (cheek: 4.41, hand: 3.94, $\cdots$) & -0.27 \\
attack (0.87) & 3.40 (people: 3.29, other: 2.83, $\cdots$) & 3.89 (enemy: 3.05, target: 3.12, $\cdots$) & -0.49 \\
plant (0.14) & 3.17 (kiss: 4.03, church: 3.73, $\cdots$) & 3.98 (tree: 4.35, garden: 4.02, $\cdots$) & -0.81 \\
welcome (0.91) & 3.39 (feedback: 2.95, comment: 2.93, $\cdots$) & 3.78 (lady: 3.66, hon: 3.82, $\cdots$) & -0.39 \\
watch (0.01) & 3.18 (weight: 3.40, intake: 3.77, $\cdots$) & 3.52 (video: 3.64, movie: 3.65, $\cdots$) & -0.34 \\
harm (0.88) & 3.32 (health: 3.44, reputation: 2.47, $\cdots$) & 3.65 (earth: 3.58, person: 3.16, $\cdots$) & -0.33 \\
meet (0.42) & 3.12 (need: 2.16, requirement: 2.02, $\cdots$) & 3.79 (people: 3.29, friend: 3.76, $\cdots$) & -0.67 \\
ride (0.19) & 3.21 (wave: 3.26, storm: 3.87, $\cdots$) & 3.88 (bike: 4.97, horse: 4.44, $\cdots$) & -0.67 \\
find (0.35) & 2.87 (way: 2.95, solution: 2.92, $\cdots$) & 4.01 (someone: 3.42, link: 3.75, $\cdots$) & -1.14 \\
gain (0.90) & 3.15 (experience: 2.13, weight: 3.40, $\cdots$) & 3.88 (pound: 4.21, muscle: 3.67, $\cdots$) & -0.73 \\
kill (0.47) & 3.39 (bacteria: 3.34, time: 2.97, $\cdots$) & 3.78 (man: 4.13, someone: 3.42, $\cdots$) & -0.39 \\
carry (0.28) & 3.10 (weight: 3.40, risk: 2.64, $\cdots$) & 3.78 (job: 3.47, work: 2.82, $\cdots$) & -0.68 \\
voice (0.54) & 2.44 (opinion: 1.92, support: 2.38, $\cdots$) & 3.39 (character: 3.02, mail: 4.07, $\cdots$) & -0.94 \\
cross (0.28) & 3.17 (mind: 2.36, boundary: 3.36, $\cdots$) & 4.00 (border: 3.68, street: 4.60, $\cdots$) & -0.83 \\
hand (0.22) & 3.05 (sentence: 3.15, power: 2.62, $\cdots$) & 3.94 (card: 4.30, key: 2.85, $\cdots$) & -0.89 \\
free (0.86) & 3.37 (time: 2.97, space: 3.47, $\cdots$) & 4.16 (slave: 3.41, prisoner: 3.54, $\cdots$) & -0.79 \\
cut (0.53) & 3.26 (cost: 2.77, corner: 4.95, $\cdots$) & 4.15 (hair: 4.43, piece: 4.16, $\cdots$) & -0.88 \\
hold (0.72) & 3.24 (breath: 3.72, that: 2.57, $\cdots$) & 4.26 (hand: 3.94, button: 4.70, $\cdots$) & -1.02 \\
waste (0.93) & 3.52 (time: 2.97, money: 3.88, $\cdots$) & 3.89 (food: 3.99, water: 4.21, $\cdots$) & -0.37 \\
send (0.10) & 3.44 (signal: 3.26, shiver: 3.78, $\cdots$) & 3.70 (email: 3.59, letter: 3.95, $\cdots$) & -0.26 \\
lose (0.98) & 3.43 (weight: 3.40, job: 3.47, $\cdots$) & 4.37 (hair: 4.43, key: 2.85, $\cdots$) & -0.94 \\
raid (0.07) & 3.28 (fund: 3.43, saving: 3.27, $\cdots$) & 4.02 (home: 3.87, house: 3.98, $\cdots$) & -0.74 \\
put (0.67) & 3.23 (effort: 2.51, pressure: 3.26, $\cdots$) & 4.28 (hand: 3.94, arm: 4.33, $\cdots$) & -1.05 \\
cost (0.15) & 3.24 (life: 3.08, job: 3.47, $\cdots$) & 3.49 (£: 3.55, much: 2.77, $\cdots$) & -0.25 \\
exchange (0.53) & 3.16 (information: 2.80, idea: 2.37, $\cdots$) & 3.93 (item: 3.65, gift: 3.55, $\cdots$) & -0.77 \\
take (0.70) & 3.18 (place: 3.08, time: 2.97, $\cdots$) & 4.17 (feed: 4.18, medication: 3.83, $\cdots$) & -0.99 \\
witness (0.03) & 3.22 (increase: 2.75, surge: 3.18, $\cdots$) & 3.24 (that: 2.57, event: 3.09, $\cdots$) & -0.02 \\
shed (0.88) & 3.21 (light: 3.77, tear: 4.24, $\cdots$) & 4.24 (hair: 4.43, coat: 4.61, $\cdots$) & -1.03 \\
piece (0.94) & 3.24 (story: 3.69, puzzle: 3.96, $\cdots$) & 4.12 (block: 3.88, quilt: 4.43, $\cdots$) & -0.87 \\
break (0.88) & 3.25 (law: 3.14, bank: 4.08, $\cdots$) & 4.35 (leg: 4.54, window: 4.67, $\cdots$) & -1.11 \\
save (0.54) & 3.30 (money: 3.88, time: 2.97, $\cdots$) & 3.81 (world: 3.04, file: 4.15, $\cdots$) & -0.52 \\
track (0.87) & 3.26 (progress: 2.78, movement: 2.69, $\cdots$) & 3.99 (copy: 3.91, vehicle: 4.01, $\cdots$) & -0.74 \\
build (0.60) & 3.08 (relationship: 2.78, business: 3.64, $\cdots$) & 4.11 (home: 3.87, house: 3.98, $\cdots$) & -1.03 \\
raise (0.81) & 3.16 (money: 3.88, awareness: 2.32, $\cdots$) & 4.28 (hand: 3.94, head: 4.04, $\cdots$) & -1.11 \\
allow (0.93) & 3.35 (user: 3.63, people: 3.29, $\cdots$) & 3.50 (run: 3.61, ourselves: 2.16, $\cdots$) & -0.15 \\
lift (0.41) & 3.12 (spirit: 2.36, ban: 3.35, $\cdots$) & 4.12 (weight: 3.40, head: 4.04, $\cdots$) & -1.00 \\
teach (0.06) & 2.79 (patience: 2.49, body: 3.81, $\cdots$) & 3.57 (child: 4.25, class: 3.33, $\cdots$) & -0.77 \\
buy (0.00) & 2.49 (time: 2.97, happiness: 2.51, $\cdots$) & 3.86 (product: 3.04, home: 3.87, $\cdots$) & -1.36 \\
\hline
All verbs (0.48) & 3.16 & 3.99 & -0.72\\
\hline
\end{tabular}
\caption{Average concreteness for metaphorical and non-metaphorical usages, and object examples for each verb.}
\label{tab:conc-table}
\end{table*}

\begin{table*}[t!]
\small
\centering
\begin{tabular}{l|l|l|c}
\hline
Verb (Metaphor rate) & Metaphorical (Object examples) & Non-metaphorical (Object examples) & Diff\\ \hline
pocket (0.28) & 3.44 (profit: 2.96, fee: 3.47, $\cdots$) & 4.00 (money: 4.12, cash: 4.17, $\cdots$) & -0.57 \\
eat (0.04) & 3.74 (world: 3.71, space: 4.05, $\cdots$) & 4.08 (food: 3.99, meal: 4.56, $\cdots$) & -0.35 \\
pull (0.36) & 3.58 (stop: 4.10, string: 3.82, $\cdots$) & 4.30 (trigger: 3.71, hair: 4.82, $\cdots$) & -0.72 \\
spell (0.65) & 3.31 (end: 3.48, trouble: 3.83, $\cdots$) & 3.86 (name: 3.54, word: 3.35, $\cdots$) & -0.55 \\
join (0.97) & 3.84 (team: 3.62, group: 3.51, $\cdots$) & 4.21 (guy: 4.39, parent: 3.47, $\cdots$) & -0.37 \\
milk (0.76) & 3.54 (system: 3.32, money: 4.12, $\cdots$) & 4.38 (cow: 4.95, goat: 4.95, $\cdots$) & -0.84 \\
express (0.99) & 3.34 (gratitude: 3.36, concern: 2.58, $\cdots$) & 4.21 (milk: 4.59, gland: 4.10, $\cdots$) & -0.87 \\
pick (0.25) & 3.55 (fight: 4.15, pace: 3.76, $\cdots$) & 4.24 (phone: 4.61, book: 4.01, $\cdots$) & -0.68 \\
make (0.87) & 3.69 (sense: 2.26, decision: 2.99, $\cdots$) & 4.39 (own: 3.17, these: 2.87, $\cdots$) & -0.70 \\
tell (0.04) & 3.67 (difference: 3.06, reality: 3.16, $\cdots$) & 3.77 (truth: 3.34, that: 3.19, $\cdots$) & -0.11 \\
view (0.35) & 3.48 (world: 3.71, thing: 3.39, $\cdots$) & 3.90 (video: 4.24, list: 3.66, $\cdots$) & -0.42 \\
kiss (0.05) & 3.96 (sky: 4.51, chance: 3.47, $\cdots$) & 4.13 (cheek: 4.77, hand: 4.08, $\cdots$) & -0.17 \\
attack (0.87) & 3.68 (people: 3.54, other: 3.18, $\cdots$) & 4.12 (enemy: 3.71, target: 3.49, $\cdots$) & -0.43 \\
plant (0.14) & 3.64 (kiss: 4.62, church: 3.82, $\cdots$) & 4.12 (tree: 4.70, garden: 4.56, $\cdots$) & -0.48 \\
welcome (0.91) & 3.70 (feedback: 3.41, comment: 3.60, $\cdots$) & 4.08 (lady: 4.17, hon: 4.03, $\cdots$) & -0.38 \\
watch (0.01) & 3.44 (weight: 3.70, intake: 3.75, $\cdots$) & 4.01 (video: 4.24, movie: 4.37, $\cdots$) & -0.57 \\
harm (0.88) & 3.65 (health: 3.71, reputation: 3.13, $\cdots$) & 3.97 (earth: 4.12, person: 3.71, $\cdots$) & -0.32 \\
meet (0.42) & 3.50 (need: 2.81, requirement: 2.69, $\cdots$) & 4.07 (people: 3.54, friend: 4.32, $\cdots$) & -0.58 \\
ride (0.19) & 3.62 (wave: 3.73, storm: 4.29, $\cdots$) & 4.06 (bike: 4.95, horse: 4.66, $\cdots$) & -0.44 \\
find (0.35) & 3.37 (way: 3.32, solution: 3.08, $\cdots$) & 4.22 (someone: 3.98, link: 4.06, $\cdots$) & -0.85 \\
gain (0.90) & 3.56 (experience: 2.96, weight: 3.70, $\cdots$) & 4.08 (pound: 4.20, muscle: 4.04, $\cdots$) & -0.53 \\
kill (0.47) & 3.72 (bacteria: 3.54, time: 3.58, $\cdots$) & 4.13 (man: 4.64, someone: 3.98, $\cdots$) & -0.41 \\
carry (0.28) & 3.56 (weight: 3.70, risk: 3.00, $\cdots$) & 3.98 (job: 3.93, work: 3.53, $\cdots$) & -0.43 \\
voice (0.54) & 3.12 (opinion: 2.70, support: 3.02, $\cdots$) & 3.81 (character: 3.86, mail: 4.15, $\cdots$) & -0.69 \\
cross (0.28) & 3.56 (mind: 3.31, boundary: 3.41, $\cdots$) & 4.16 (border: 3.85, street: 4.30, $\cdots$) & -0.60 \\
hand (0.22) & 3.45 (sentence: 3.40, power: 3.40, $\cdots$) & 4.14 (card: 4.41, key: 3.46, $\cdots$) & -0.70 \\
free (0.86) & 3.74 (time: 3.58, space: 4.05, $\cdots$) & 4.32 (slave: 3.98, prisoner: 4.03, $\cdots$) & -0.58 \\
cut (0.53) & 3.61 (cost: 2.96, corner: 4.57, $\cdots$) & 4.28 (hair: 4.82, piece: 4.32, $\cdots$) & -0.67 \\
hold (0.72) & 3.59 (breath: 4.01, that: 3.19, $\cdots$) & 4.39 (hand: 4.08, button: 4.49, $\cdots$) & -0.79 \\
waste (0.93) & 3.77 (time: 3.58, money: 4.12, $\cdots$) & 4.04 (food: 3.99, water: 4.18, $\cdots$) & -0.27 \\
send (0.10) & 3.83 (signal: 3.62, shiver: 4.44, $\cdots$) & 3.89 (email: 3.83, letter: 4.07, $\cdots$) & -0.07 \\
lose (0.98) & 3.77 (weight: 3.70, job: 3.93, $\cdots$) & 4.47 (hair: 4.82, key: 3.46, $\cdots$) & -0.71 \\
raid (0.07) & 3.54 (fund: 3.35, saving: 3.80, $\cdots$) & 3.99 (home: 4.15, house: 4.23, $\cdots$) & -0.44 \\
put (0.67) & 3.63 (effort: 3.01, pressure: 3.48, $\cdots$) & 4.37 (hand: 4.08, arm: 4.11, $\cdots$) & -0.75 \\
cost (0.15) & 3.69 (life: 3.90, job: 3.93, $\cdots$) & 3.76 (£: 3.38, much: 3.35, $\cdots$) & -0.07 \\
exchange (0.53) & 3.60 (information: 3.16, idea: 3.13, $\cdots$) & 4.10 (item: 3.83, gift: 4.06, $\cdots$) & -0.50 \\
take (0.70) & 3.60 (place: 3.51, time: 3.58, $\cdots$) & 4.31 (feed: 4.22, medication: 4.08, $\cdots$) & -0.71 \\
witness (0.03) & 3.43 (increase: 2.96, surge: 3.60, $\cdots$) & 3.78 (that: 3.19, event: 3.66, $\cdots$) & -0.35 \\
shed (0.88) & 3.59 (light: 4.10, tear: 4.34, $\cdots$) & 4.37 (hair: 4.82, coat: 4.68, $\cdots$) & -0.78 \\
piece (0.94) & 3.70 (story: 4.13, puzzle: 4.32, $\cdots$) & 4.24 (block: 4.07, quilt: 4.60, $\cdots$) & -0.54 \\
break (0.88) & 3.66 (law: 3.33, bank: 3.96, $\cdots$) & 4.42 (leg: 4.44, window: 4.57, $\cdots$) & -0.76 \\
save (0.54) & 3.67 (money: 4.12, time: 3.58, $\cdots$) & 4.08 (world: 3.71, file: 4.13, $\cdots$) & -0.41 \\
track (0.87) & 3.57 (progress: 3.26, movement: 3.29, $\cdots$) & 4.27 (copy: 4.04, vehicle: 4.23, $\cdots$) & -0.70 \\
build (0.60) & 3.49 (relationship: 3.54, business: 3.71, $\cdots$) & 4.24 (home: 4.15, house: 4.23, $\cdots$) & -0.75 \\
raise (0.81) & 3.54 (money: 4.12, awareness: 2.98, $\cdots$) & 4.37 (hand: 4.08, head: 4.13, $\cdots$) & -0.84 \\
allow (0.93) & 3.69 (user: 3.91, people: 3.54, $\cdots$) & 3.92 (run: 3.82, ourselves: 2.85, $\cdots$) & -0.22 \\
lift (0.41) & 3.54 (spirit: 3.35, ban: 3.64, $\cdots$) & 4.26 (weight: 3.70, head: 4.13, $\cdots$) & -0.72 \\
teach (0.06) & 3.43 (patience: 3.45, body: 4.15, $\cdots$) & 3.80 (child: 4.36, class: 3.51, $\cdots$) & -0.37 \\
buy (0.00) & 3.21 (time: 3.58, happiness: 3.56, $\cdots$) & 3.80 (product: 3.18, home: 4.15, $\cdots$) & -0.58 \\
\hline
All verbs (0.48) & 3.57 & 4.18 & -0.56\\
\hline
\end{tabular}
\caption{Average imageability for metaphorical and non-metaphorical usages, and object examples for each verb.}
\label{tab:img-table}
\end{table*}

\begin{table*}[t!]
\small
\centering
\begin{tabular}{l|l|l|c}
\hline
Verb (Metaphor rate) & Metaphorical (Object examples) & Non-metaphorical (Object examples) & Diff\\ \hline
pocket (0.28) & 3.99 (profit: 4.60, fee: 4.43, $\cdots$) & 4.21 (money: 4.67, cash: 4.50, $\cdots$) & -0.23 \\
eat (0.04) & 3.81 (space: 4.29, cost: 4.43, $\cdots$) & 4.07 (meal: 4.43, meat: 4.50, $\cdots$) & -0.26 \\
pull (0.36) & 3.61 (stop: 4.71, string: 4.14, $\cdots$) & 3.89 (trigger: 3.14, hair: 4.71, $\cdots$) & -0.29 \\
spell (0.65) & 3.60 (end: 4.86, trouble: 4.00, $\cdots$) & 4.15 (name: 4.71, word: 4.71, $\cdots$) & -0.55 \\
join (0.97) & 3.76 (team: 4.71, force: 4.29, $\cdots$) & 3.79 (guy: 4.57, parent: 4.57, $\cdots$) & -0.03 \\
milk (0.76) & 3.99 (system: 4.20, money: 4.67, $\cdots$) & 4.23 (cow: 4.43, goat: 4.29, $\cdots$) & -0.25 \\
express (0.99) & 3.19 (gratitude: 2.83, concern: 3.29, $\cdots$) & 4.08 (milk: 4.50, gland: 2.33, $\cdots$) & -0.89 \\
pick (0.25) & 3.63 (fight: 4.43, pace: 3.57, $\cdots$) & 3.91 (item: 4.50, book: 4.71, $\cdots$) & -0.29 \\
make (0.87) & 3.64 (sense: 4.29, decision: 3.71, $\cdots$) & 4.11 (own: 4.71, these: 4.71, $\cdots$) & -0.47 \\
tell (0.04) & 3.68 (difference: 3.43, reality: 3.43, $\cdots$) & 3.86 (truth: 4.60, that: 4.86, $\cdots$) & -0.18 \\
view (0.35) & 3.46 (thing: 4.86, themselves: 4.50, $\cdots$) & 3.70 (list: 4.33, information: 3.71, $\cdots$) & -0.24 \\
kiss (0.05) & 4.24 (chance: 4.17, dream: 4.50, $\cdots$) & 4.15 (cheek: 4.43, hand: 4.71, $\cdots$) & \textit{{\scriptsize +}0.09} \\
attack (0.87) & 3.77 (people: 4.71, other: 4.71, $\cdots$) & 3.60 (enemy: 4.33, target: 4.00, $\cdots$) &  \textit{{\scriptsize +}0.16} \\
plant (0.14) & 3.89 (kiss: 4.40, idea: 4.71, $\cdots$) & 4.14 (tree: 4.50, plant: 4.57, $\cdots$) & -0.24 \\
welcome (0.91) & 3.73 (feedback: 3.43, comment: 4.00, $\cdots$) & 3.45 (president: 3.57, lady: 4.71, $\cdots$) &  \textit{{\scriptsize +}0.28} \\
watch (0.01) & 3.19 (weight: 4.00, intake: 3.29, $\cdots$) & 3.92 (movie: 4.43, show: 4.57, $\cdots$) & -0.73 \\
harm (0.88) & 3.68 (reputation: 2.60, economy: 3.29, $\cdots$) & 3.83 (earth: 4.57, person: 4.43, $\cdots$) & -0.15 \\
meet (0.42) & 3.64 (need: 4.67, requirement: 3.86, $\cdots$) & 3.66 (people: 4.71, friend: 4.71, $\cdots$) & -0.02 \\
ride (0.19) & 3.90 (wave: 4.17, momentum: 2.83, $\cdots$) & 4.18 (bike: 4.57, horse: 4.57, $\cdots$) & -0.29 \\
find (0.35) & 3.44 (way: 4.71, solution: 3.71, $\cdots$) & 3.91 (someone: 4.60, link: 4.29, $\cdots$) & -0.47 \\
gain (0.90) & 3.75 (experience: 3.71, weight: 4.00, $\cdots$) & 3.98 (muscle: 3.86, lb: 4.17, $\cdots$) & -0.23 \\
kill (0.47) & 3.76 (bacteria: 3.00, time: 5.00, $\cdots$) & 3.74 (man: 4.57, someone: 4.60, $\cdots$) &  \textit{{\scriptsize +}0.02} \\
carry (0.28) & 3.53 (weight: 4.00, risk: 3.60, $\cdots$) & 3.69 (job: 4.86, work: 4.29, $\cdots$) & -0.16 \\
voice (0.54) & 3.52 (opinion: 3.43, support: 3.86, $\cdots$) & 4.14 (character: 3.71, mail: 4.29, $\cdots$) & -0.62 \\
cross (0.28) & 3.80 (mind: 4.43, boundary: 2.67, $\cdots$) & 3.86 (border: 3.71, arm: 4.83, $\cdots$) & -0.05 \\
hand (0.22) & 3.54 (sentence: 3.67, power: 4.29, $\cdots$) & 3.96 (card: 4.57, key: 4.50, $\cdots$) & -0.42 \\
free (0.86) & 3.86 (time: 5.00, space: 4.29, $\cdots$) & 3.99 (slave: 4.00, prisoner: 3.29, $\cdots$) & -0.12 \\
cut (0.53) & 3.62 (cost: 4.43, corner: 3.83, $\cdots$) & 3.93 (hair: 4.71, piece: 4.17, $\cdots$) & -0.31 \\
hold (0.72) & 3.62 (breath: 3.80, that: 4.86, $\cdots$) & 3.99 (hand: 4.71, button: 4.29, $\cdots$) & -0.36 \\
waste (0.93) & 3.86 (time: 5.00, money: 4.67, $\cdots$) & 4.16 (water: 4.71, organism: 3.29, $\cdots$) & -0.30 \\
send (0.10) & 3.76 (signal: 3.50, prayer: 3.83, $\cdots$) & 3.99 (letter: 4.33, information: 3.71, $\cdots$) & -0.23 \\
lose (0.98) & 3.72 (weight: 4.00, job: 4.86, $\cdots$) & 3.95 (hair: 4.71, key: 4.50, $\cdots$) & -0.22 \\
raid (0.07) & 3.73 (saving: 4.71, wealth: 3.71, $\cdots$) & 4.21 (closet: 3.86, place: 4.67, $\cdots$) & -0.47 \\
put (0.67) & 3.67 (effort: 3.86, pressure: 3.86, $\cdots$) & 3.89 (hand: 4.71, arm: 4.83, $\cdots$) & -0.23 \\
cost (0.15) & 3.79 (life: 4.57, job: 4.86, $\cdots$) & 3.94 (much: 4.57, money: 4.67, $\cdots$) & -0.15 \\
exchange (0.53) & 3.88 (information: 3.71, idea: 4.71, $\cdots$) & 4.06 (item: 4.50, money: 4.67, $\cdots$) & -0.18 \\
take (0.70) & 3.60 (place: 4.67, time: 5.00, $\cdots$) & 3.95 (feed: 4.14, medication: 2.29, $\cdots$) & -0.35 \\
witness (0.03) & 3.18 (increase: 4.29, surge: 3.50, $\cdots$) & 3.92 (that: 4.86, event: 4.29, $\cdots$) & -0.74 \\
shed (0.88) & 3.82 (light: 4.50, tear: 4.33, $\cdots$) & 4.08 (hair: 4.71, coat: 4.40, $\cdots$) & -0.26 \\
piece (0.94) & 4.07 (story: 4.57, thing: 4.86, $\cdots$) & 4.20 (block: 4.43, back: 4.71, $\cdots$) & -0.13 \\
break (0.88) & 3.68 (law: 4.43, rule: 4.43, $\cdots$) & 4.00 (leg: 4.71, window: 4.43, $\cdots$) & -0.31 \\
save (0.54) & 3.77 (money: 4.67, time: 5.00, $\cdots$) & 3.87 (file: 4.43, site: 4.33, $\cdots$) & -0.10 \\
track (0.87) & 3.64 (progress: 3.71, movement: 3.71, $\cdots$) & 3.87 (copy: 4.57, vehicle: 3.43, $\cdots$) & -0.23 \\
build (0.60) & 3.55 (trust: 4.50, confidence: 3.33, $\cdots$) & 3.71 (plant: 4.57, facility: 2.67, $\cdots$) & -0.16 \\
raise (0.81) & 3.70 (money: 4.67, awareness: 3.17, $\cdots$) & 4.06 (hand: 4.71, head: 4.71, $\cdots$) & -0.36 \\
allow (0.93) & 3.34 (people: 4.71, time: 5.00, $\cdots$) & 3.85 (run: 5.00, ourselves: 4.00, $\cdots$) & -0.50 \\
lift (0.41) & 3.65 (ban: 4.00, mood: 4.20, $\cdots$) & 4.03 (weight: 4.00, head: 4.71, $\cdots$) & -0.37 \\
teach (0.06) & 3.41 (patience: 3.29, body: 4.57, $\cdots$) & 3.88 (child: 4.71, class: 4.29, $\cdots$) & -0.47 \\
buy (0.00) & 3.31 (time: 5.00, happiness: 4.14, $\cdots$) & 4.11 (ticket: 4.43, book: 4.71, $\cdots$) & -0.80 \\
\hline
All verbs (0.48) & 3.63 & 3.87 & -0.26\\
\hline
\end{tabular}
\caption{Average familiarity for metaphorical and non-metaphorical usages, and object examples for each verb.}
\label{tab:fam-table}
\end{table*}
\end{document}